\documentclass[preprint,review,12pt,authoryear]{elsarticle}

\usepackage{amssymb}
\usepackage{amsmath}

\usepackage{diagbox}

\journal{Neural Networks}

\begin{document}

\begin{frontmatter}



\title{Adaptive Point Transformer}


\author[inst1]{Alessandro Baiocchi\corref{cor1}}

\cortext[cor1]{Email: alessandro.baiocchi@uniroma1.it}

\affiliation[inst1]{organization={Sapienza University of Rome, Department of Computer, Control and Management Engineering},
            addressline={Via Ariosto 25}, 
            city={Rome},
            postcode={00185},
            country={Italy}}

\affiliation[inst3]{organization={Sapienza University of Rome, Department of Computer Science},
            addressline={via Salaria, 113}, 
            city={Rome},
            postcode={00198},
            country={Italy}}

\affiliation[inst4]{organization={Leonardo Labs},
            addressline={Piazza Monte Grappa, 4}, 
            city={Rome},
            postcode={00195},
            country={Italy}}

\affiliation[inst2]{organization={Sapienza University of Rome, Department of Electronics and Telecommunication Engineering},
            addressline={via Eudossiana, 18}, 
            city={Rome},
            postcode={00184},
            country={Italy}}
            
\author[inst3]{Indro Spinelli}
\author[inst4]{Alessandro Nicolosi}
\author[inst2]{Simone Scardapane}

\begin{abstract}
The recent surge in 3D data acquisition has spurred the development of geometric deep learning models for point cloud processing, boosted by the remarkable success of transformers in natural language processing. While point cloud transformers (PTs) have achieved impressive results recently, their quadratic scaling with respect to the point cloud size poses a significant scalability challenge for real-world applications. To address this issue, we propose the Adaptive Point Cloud Transformer (AdaPT), a standard PT model augmented by an adaptive token selection mechanism. AdaPT dynamically reduces the number of tokens during inference, enabling efficient processing of large point clouds. Furthermore, we introduce a budget mechanism to flexibly adjust the computational cost of the model at inference time without the need for retraining or fine-tuning separate models. Our extensive experimental evaluation on point cloud classification tasks demonstrates that AdaPT significantly reduces computational complexity while maintaining competitive accuracy compared to standard PTs. The code for AdaPT is made publicly available.
\end{abstract}




\end{frontmatter}


\section{Introduction}
\label{sec:intro}

The increasing ubiquity of point cloud data across diverse domains, from autonomous navigation to environmental monitoring, has spurred significant advancements in point cloud representation learning \citep{camuffo2022recent}. Unlike traditional two-dimensional data, point clouds exhibit unique characteristics, including their inherent invariance under the permutation of their constituent points. These properties pose inherent challenges for conventional neural network architectures, necessitating the development of tailored methodologies specifically designed for point cloud processing.

Early attempts to tackle point cloud analysis relied on localized message-passing approaches inspired by graph neural networks \citep{kipf2017semi}, as exemplified by PointNet \citep{qi2017pointnet} and its extensions \citep{qi2017pointnet++,guo2020deep}. However, these methods often suffer from limited receptive field sizes, particularly when dealing with large-scale point clouds.

In recent years, the adoption of transformers \citep{vaswani2023attention}, initially introduced for natural language processing, has emerged as a promising alternative for point cloud representation learning \citep{Guo_2021}. Transformers excel in handling unordered data through their attention mechanism, which operates on tokenized representations of the input data. This tokenization enables transformers to readily adapt to diverse data types, treating each token as an independent entity, regardless of its origin. After the tokenization, the self-attention mechanism is crucial in preserving the permutation invariance of point clouds. By operating on an unordered set of tokens, self-attention allows transformers to learn relationships between points without explicitly considering their order. However, the quadratic computational complexity of self-attention poses a significant challenge for scaling transformers to large point clouds \citep{hackel2017isprs}. Several approaches have been proposed to mitigate this issue. One strategy involves using a fixed number of tokens by grouping multiple points into single tokens \citep{Guo_2021}.  This method offers computational efficiency but sacrifices accuracy due to the aggregation of heterogeneous points. Another approach involves knowledge distillation, where a large, pre-trained model (teacher) is used to train a smaller, more efficient model (student) \citep{zhang2022pointdistiller,yao20223d}. This approach can reduce inference time at the cost of reduced accuracy. 

Token selection, another technique for reducing token count, involves selectively eliminating tokens based on a relevance criteria. This method has been successfully employed in computer vision applications, such as AdaViT \citep{meng2021adavit} and DynamicViT \citep{rao2021dynamicvit}, which dynamically assess token importance for the main tasks. Yang et al. \citep{Yang2019modeling} adapted this framework for point cloud preprocessing. However, these methods require users to pre-define a token budget, limiting flexibility.

To address these shortcomings, we propose the Adaptive Point Cloud Transformer (AdaPT), a novel model that seamlessly integrates adaptive downsampling and dynamic resource allocation. AdaPT adaptively selects the number of tokens to retain based on an input parameter specified at inference time, enabling a flexible trade-off between computational resources and prediction accuracy. AdaPT's ability to adjust its computational budget based on available resources makes it well-suited for real-world applications where resource constraints necessitate efficient processing with variable inference time, especially when dealing with variable-sized point clouds. The proposed AdaPT model represents a significant step in developing scalable and adaptable point cloud representation learning methods.

\subsubsection*{Organization of the paper}

After introducing AdaPT in Section \ref{sec:adapt}, we demonstrate its effectiveness through extensive experiments in Section \ref{sec:experiments}. In Section \ref{sec:ablation} we provide an in-depth analysis of AdaPT's performance and several ablation studies to validate its performance in handling large-scale point clouds with varying computational budgets. We conclude in \ref{sec:conclusions} with some final remarks and possible future directions.

\section{Related Works}
\label{sec:relworks}

In this section, we present related works on two aspects: Point Cloud processing and Transformer-based models.

\subsection{Point Cloud Representation and Processing}
\label{sec:relworks:pcrepr}
PointNet is considered the pioneering work in point cloud learning \citep{qi2017pointnet}, followed by PointNet++ \citep{qi2017pointnet++}. These models directly process point cloud data and learn a representation of the input using a feed-forward network, which is applied independently to each point. The information from all points is then aggregated in a permutation-invariant manner. PointNet++ enhances this approach by introducing local features obtained by aggregating information from neighboring points through ball query grouping and a hierarchical PointNet structure.

Several other works have defined a convolution operator for point clouds. These models typically leverage voxelization or octrees. The former approach converts point cloud data into a regular three-dimensional voxel structure, facilitating standard convolution. The latter approach uses a tree structure combined with varying sizes of voxels to capture different densities within the point cloud. Notable examples of these models include SparseConvNet \citep{graham20173d} and OctNet \citep{riegler2017octnet}, which exploit the natural sparsity of input point clouds. However, these techniques are limited by the local receptive field of the convolution operator, which may be insufficient for large-scale point clouds.

To address this limitation, \citet{wang2019dynamic} introduced the EdgeConv operator, which enables dynamic convolution over the point cloud and captures both global and local information. Despite these advancements, convolutional approaches still underperform compared to attention-based counterparts \citep{Guo_2021}, a topic we discuss in the following subsection.

\subsection{Transformer Architectures}
\label{sec:relworks:transfarch}

Transformer models, introduced by \citet{vaswani2023attention}, were initially designed for natural language processing (NLP) tasks, exploiting the attention mechanism. These models operate on tokens derived from the input through a tokenization operation. The tokens are processed to obtain a hidden representation, typically by a task-specific head, to yield the final prediction.

The high parallelization possible in attention calculations has allowed transformer models to benefit significantly from recent GPU performance improvements, enabling larger models' training \citep{narayanan2021efficient}.

Following their introduction in NLP, the general nature of their architecture has allowed transformers to be employed across various fields, achieving outstanding performance \citep{islam2024survey}. Typically, the only adjustment needed for a transformer to work with different data types is to replace the initial tokenization module.

In image processing, for example, Vision Transformers (ViTs) have replaced Convolutional Neural Networks as the state-of-the-art for various tasks, such as image classification and image segmentation \citep{dosovitskiy2021image,thisanke2023semantic}.

Graph Transformers (GTs) \citep{mueller2023attending}, a family of architectures based on transformers, have been developed to operate on graph-structured data, leveraging the desirable properties of the attention operation.

Transformer-based models have recently emerged in point cloud data processing \citep{lu2022transformers}. In particular, \citet{Guo_2021} and \citet{zhao2021point} simultaneously introduced a transformer-based model for point cloud data. Transformers inherently possess desirable properties for point cloud deep learning, as they naturally maintain permutation invariance among the tokens, treating them as an unordered set.

Due to the inherently quadratic nature of the attention operation \citep{dumankeles2023attention}, scaling the number of tokens used in transformers is challenging. One solution involves developing dynamic models capable of subsampling only the most relevant tokens in the input.

In the field of image processing, models such as AdaViT \citep{meng2021adavit} and DynamicViT \citep{rao2021dynamicvit} have been developed, proposing methods for selecting and removing unimportant tokens during the calculations.

In the context of point cloud data, recent works, such as that by \citet{Yang2019modeling}, employ strategies analogous to those used in ViTs to subsample less relevant tokens.

Existing models have successfully enhanced the performance and interpretability of their base models, albeit at the cost of some accuracy. These models, applicable to images and point clouds, necessitate the pre-determination of a computational budget, establishing a fixed trade-off at inference time. The contribution we introduce in the subsequent section offers a significant advancement over these models. Our approach allows for an adaptive computational budget at inference time, enabling a dynamic trade-off between accuracy and computational complexity. Importantly, this adaptability is achieved without the need to re-train the model.

\section{Adaptive Point Transformer}
\label{sec:adapt}
In this work, we present the Adaptive Point Transformer (AdaPT), a model capable of classifying point cloud data using a flexible computational budget at inference time. 
To achieve this, we pair a series of transformer layers with a set of token selection modules. 
These modules subsample, in a differentiable manner, the tokens used in the point cloud transformer, reducing the computational effort needed in subsequent layers.

\begin{figure}[htbp]
    \centering
    \includegraphics[width=0.8\textwidth]{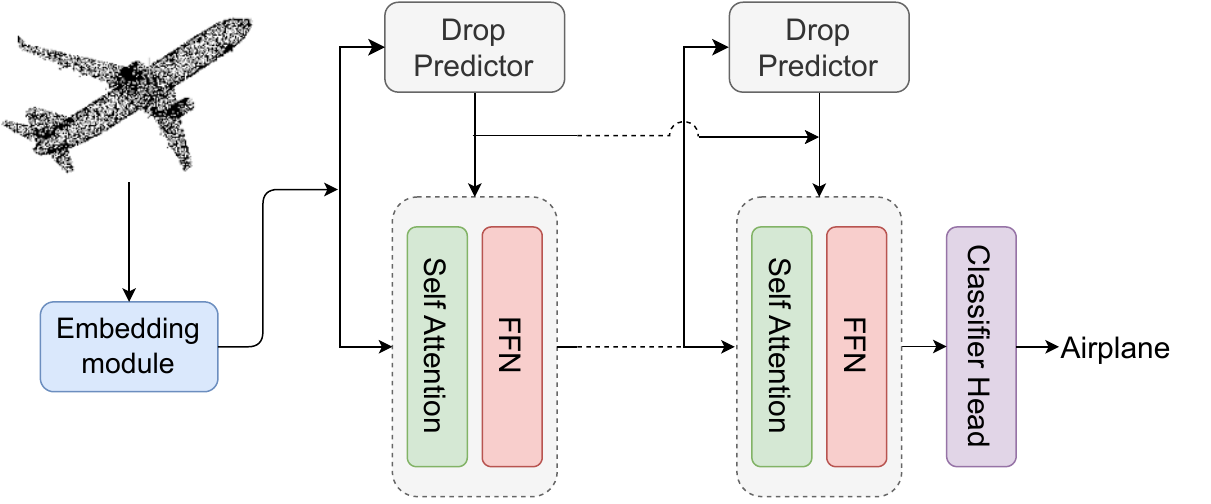}
    \caption{Architecture of the AdaPT model. The embedding module consists of an Absolute-Relative positional embedding \citep{Yang2019modeling}. The transformer blocks are paired with the proposed \textit{drop predictors}. Finally, the representation is processed by a classifier head which outputs a prediction.}
    \label{fig:fullnet}
\end{figure}

The amount of tokens sampled at each layer is not fixed. This allows for the computational budget available to the model to be decided at inference time for each data point.
In the following sections, we explain the components of the model, which are shown schematically in Fig. \ref{fig:fullnet}.

\subsection{Point Cloud Transformer}
\label{sec:PCT}

We want to build a transformer-based architecture in order to be able to drop less relevant tokens in a similar manner to \citet{rao2021dynamicvit}. We need our point cloud data to be embedded in tokens to employ a transformer model. 
We do this through the Absolute-Relative Positional Embedding (ARPE) module presented in \citet{Yang2019modeling}. 
This module embeds each point into a token utilizing information from the point itself and its nearest neighbors.
Let $x\in\mathbb{R}^{N\times F}$ be a point cloud made of $N$ points, each with $f$ features. 
Let $k_i\in\mathbb{R}^{k\times F}$ be the set of $k$ nearest points to a point $x_i\in x$.
The embedding is done as follows:
\begin{equation*}
    \text{ARPE}(x_i) = \gamma(\max\{h(x^\prime) \mid x^\prime \in k_i\})
\end{equation*}
where both $\gamma$ and $h$ are Multi-Layer Perceptrons (MLPs) with group normalization and ELU activation \citep{clevert2016fast}. 

The tokens are then processed by a series of transformer layers employing multi-head attention (MHA), layer normalization (LN), and a MLP composed of two linear layers with GELU activations:
\begin{gather*}
    z_0 =  \text{ARPE}(x_i)\\
    z_\ell = \text{MLP(LN(MHA(LN}(z_{\ell-1}))))
\end{gather*}
The final token representations are averaged, and the result is passed through a classification head made of a MLP composed of two linear layers with GELU activations, which outputs the network's prediction:

\begin{equation*}
    z^\prime_f = \frac{1}{n}\sum_{i=0}^n z_{f,i} \qquad\qquad y = \text{MLP}(z^\prime_f)
\end{equation*}
where $z^\prime_f$ is the average computed over the final tokens.



\subsection{Drop predictor module}
\label{sec:droppred}

\begin{figure}[htb]
    \begin{minipage}[t]{.48\textwidth}
        \centering
        \includegraphics[width=\textwidth]{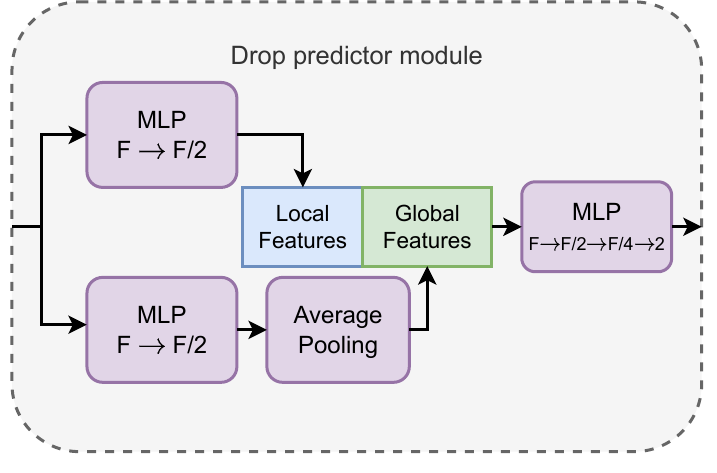}
        \caption{Scheme of the drop predictor module architecture.}\label{fig:dropscheme}
    \end{minipage}
    \hfill
    \begin{minipage}[t]{.48\textwidth}
        \centering
        \includegraphics[width=\textwidth]{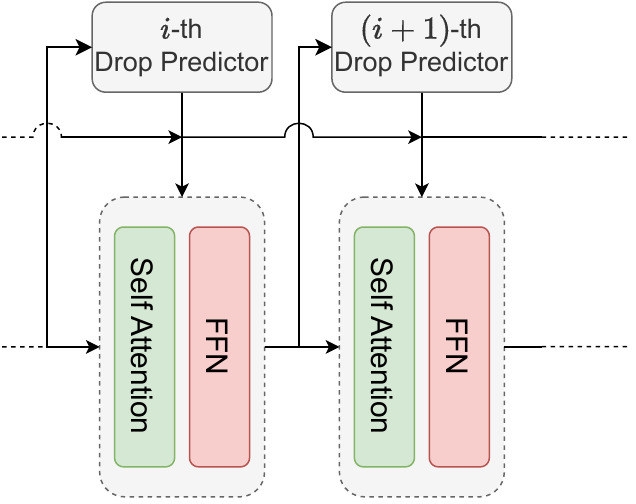}
        \caption{Drop predictor modules usage in the PCT.}\label{fig:dropfull}
    \end{minipage}  
    \label{fig:dropmodule}
\end{figure}

The main aim of the AdaPT model is to reduce the computational effort during inference by subsampling the tokens. 
This is done through modules responsible for the token selection, which are called ``drop predictors'', whose architecture is shown in Fig. \ref{fig:dropscheme}. 
These modules take into consideration both local (relative to the single token) and global (relative to all tokens) features.
Let $x\in\mathbb{R}^{n\times F}$ be the input tokens to a drop predictor layer. We first compute what we call local and global features as:

\begin{equation*}
    z_{local} = \text{MLP}(x) \in \mathbb{R}^{n \times F^\prime} \qquad z_{global} = \text{Agg(MLP}(x))\in \mathbb{R}^{F^\prime}
\end{equation*}
where Agg is an aggregation operation, e.g. average pooling, and $F^\prime$ is an arbitrary dimension, in our case, $F^\prime = F/2$. The global features are repeated $n$ times, obtaining a new $z_{global} \in \mathbb{R}^{n\times f\prime}$. These features are concatenated and fed to an additional MLP that predicts the probabilities of keeping the tokens:

\begin{equation*}
    z = [z_{local}, z_{global}] \in \mathbb{R}^{n\times F}\qquad \qquad \pi = \text{Softmax}(\text{MLP}(z)) \in \mathbb{R}^{n\times2}
\end{equation*}
where $\pi_{i,0}$ denotes the probability of eliminating the $i$-th token, while $\pi_{i,1}$ denotes the probability of keeping it.

We also want the model to learn to select the least relevant tokens for a given task and eliminate them.
For end-to-end training to be possible, we need to sample the tokens in a differentiable manner. 
This is done by employing the Gumbel-Softmax (GS) estimator \citep{jang2017categorical} to differentiably sample from the probability distributions $\pi_j \in \mathbb{R}^2; j=1,...,N$:

\begin{equation}
    \text{GS}(\pi_j, \tau)_{i} = \frac{\exp((\log(\pi_{j,i})+g_{j,i})/\tau)}{\sum_{k=0,1}\exp((\log(\pi_{j,k})+g_{j,k})/\tau)} \quad i=0,1 ; j=0,...,N
\end{equation}
where $g_{i,j}$ are i.i.d. samples drawn from Gumbel(0,1)\footnote{The Gumbel distribution samples can be drawn by computing $g = -\log(-\log(u))$, where $u$ is drawn from \text{Uniform(0,1)}.}, and $\tau$ is the temperature parameter, which is set to $\tau=1$ in our experiments. 

To obtain a binary output an argmax is applied during the forward pass, while straight-through estimation \citep{jang2017categorical} is used for the computation of the gradients. 
We obtain the binary mask $D \in \{0,1\}^N$ we use to index the tokens as:
\begin{equation}
    D_j = \text{argmax}(\text{GS}(\pi_j, \tau))_{1} \in \{0,1\} \qquad j = 1, ..., N
\end{equation}

We use only the values corresponding to keeping the tokens so that $D$ can be used directly as a boolean mask to index the tokens. We add drop predictor modules to a subset of the transformer blocks in the PCT, as shown in Fig. \ref{fig:dropfull}.

\begin{figure}[htbp]
    \centering
    \includegraphics[width=0.65\textwidth]{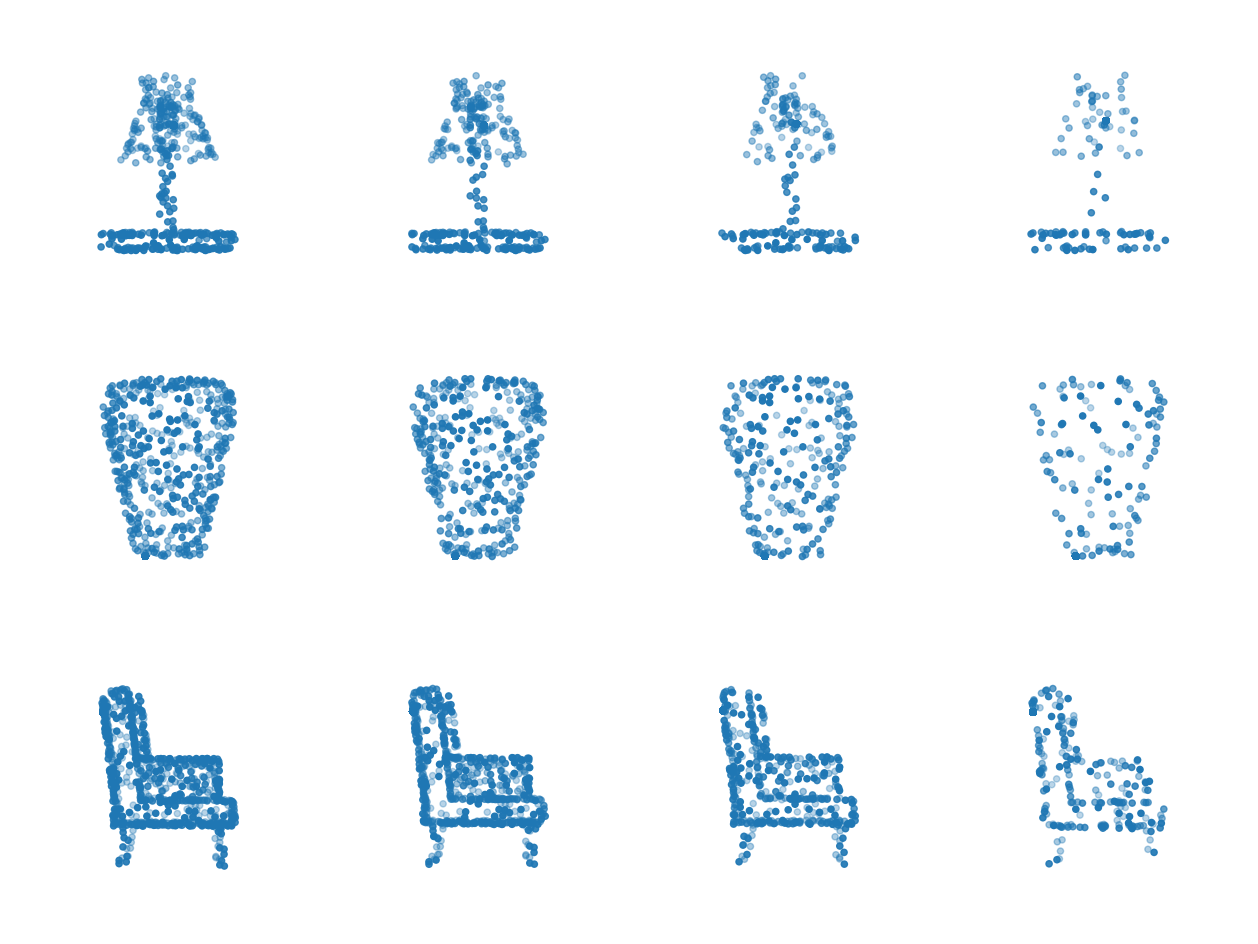}
    \caption{Visualization of the kept tokens along the model's layers for a few representative examples.}
    \label{fig:visualization}
\end{figure}

During the training phase, the tokens are not eliminated; they are instead masked in the attention, effectively preventing them from contributing to the prediction in successive steps.
At inference time, the tokens are completely removed, boosting the computational efficiency in the successive layers. 
A demonstration of the effect of this token elimination is shown in Fig. \ref{fig:visualization}.
Each drop predictor module also passes its decision on which tokens to eliminate to the next layers in such a way as to ensure that, during the training phase, once a token is masked, it stays masked for all of the remaining computations.

Additionally, we add a regularization term to the loss to incentivize the model to drop tokens. We describe the regularization term for a single budget here, and its extension to a flexible budget in the next section. Denote by $t_i$ the target ratio of dropped tokens for the $i$-th drop predictor. The regularization term we consider is the following:
\begin{equation}
\label{eq:regterm}
    \mathcal{L}_{drop} = \sum^{p}_{i=0}\frac{(d_i - t_i)^2}{p}
\end{equation}
where $p$ is the total number of drop predictors, while $d_i$ and $t_i$ are the fraction of token dropped and the target drop ratio for the $i$-th drop predictor.

This regularization term is added to a standard cross entropy loss term $\mathcal{L}_{CE}$.
The complete loss function is then:
\begin{equation}
    \mathcal{L} = \mathcal{L}_{CE} + \alpha\mathcal{L}_{drop}
\end{equation}
where $\alpha$ is a parameter controlling the strength of the regularization.

\subsection{Flexible budget}
\label{sec:budget}

\begin{figure}[htbp]
    \centering
    \includegraphics[width=0.8\textwidth]{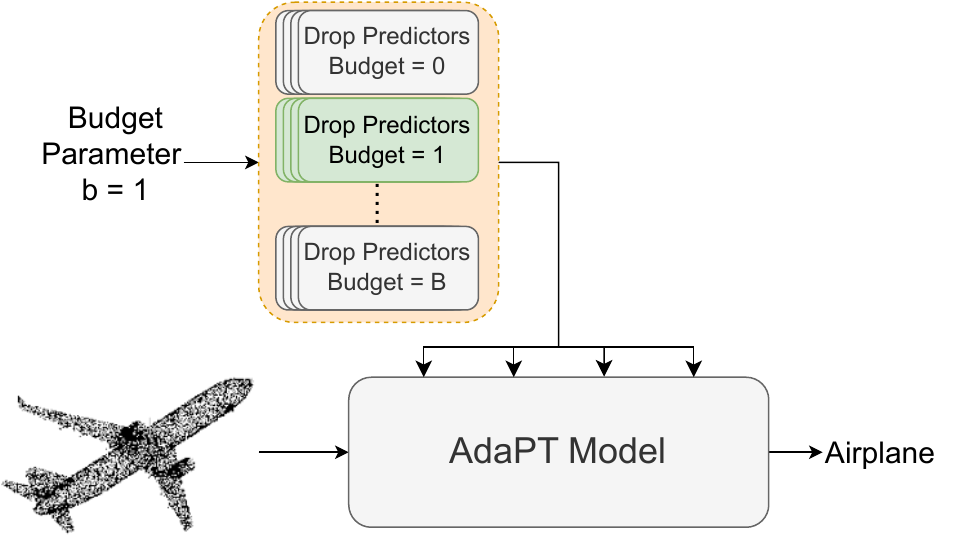}
    \caption{The budget parameter selects the set of drop predictors to be used in the classification of a point cloud.
    This parameter also selects a specific regularization term that is used to train the corresponding drop predictor set.}
    \label{fig:budget}
\end{figure}

Finally, we want our trained model to be able to operate with multiple computational budgets, which are to be selected by the user at inference time. To achieve this, we add to the model multiple sets of drop predictors in parallel, one for each desired budget. Note that each drop predictor layer (as described in Section \ref{sec:droppred}) is extremely small, thus adding a low overhead of parameters to the complete model. These $B$ sets of drop predictors are all trained simultaneously while sharing the same backbone model. This is shown schematically in Fig. \ref{fig:budget}.

Let $b \in \{0, ..., B\}$ be an index variable denoting increasingly higher budget values set by the user. We use $t^b_i$ to denote the corresponding $b$-th desired budget at the $i$-th layer. In addition, denote by $d^b_i$ the ratio of dropped tokens at layer $i$ if we use the $b$-th drop predictor module. The regularization term in \eqref{eq:regterm} becomes:
\begin{equation}
    \mathcal{L}_{drop} = \sum_{b=1}^B \sum^{p}_{i=0}\frac{(d_i^b - t^b_i)^2}{p}
\end{equation}
where we want all budgets to be simultaneously satisfied depending on the index $b$. Computing this regularization term is unfeasible because it would require $B$ forward passes for each input point cloud (one for each budget). To this end, during training a random budget is extracted for each batch of data, and the corresponding regularization term and set of drop predictor modules are used for that batch.

During inference, the budget is an additional input required by the model, which selects the appropriate set of drop predictor modules.
By selecting a set of modules trained with a certain regularization term, this parameter selects the ratio of tokens that will be eliminated during inference, hence selecting the computational budget for that prediction.

\section{Experimental section}
\label{sec:experiments}

In this section, we compare our model to two other transformer-based architectures for point cloud classification, presented in \citet{Guo_2021} and \citet{zhao2021point}. 
The dataset used for these comparisons is ModelNet40 \citep{wu20153d}.
To allow for a fair comparison of the results, we implemented the models used in these works into our pipeline, training and testing the models on the same split of the dataset and with the same data preprocessing.

\subsection{Regularization target and Budget parameter}
We set the target used for the regularization presented in \eqref{eq:regterm} using two hyperparameters: the number of layers with a drop predictor associated, $\ell$, and the desired ratio of dropped tokens to initial tokens after the last layer, $\rho$.
Once those parameters are fixed, the target ratio for each layer is computed as:

\begin{equation}
    t_i = \frac{i}{\ell}\cdot\rho \qquad i = 1, ..., \ell
\end{equation}

Additionally, we need to set the target for different values of the budget parameter. 
We choose to have the target depend linearly on the value of the budget parameter, so that the only hyperparameter to be set is the number of different budgets available, $B$.
The targets associated to the different budgets are then computed as follows:
\begin{equation}
    t_i^b = \frac{B-b}{B-1}\cdot t_i = \frac{(B-b)\cdot i}{(B-1)\cdot\ell}\cdot\rho \qquad b = 1, ..., B \quad i = 1, ..., \ell 
\end{equation}

We set $\ell = 4$, $\rho = 0.8$, and $B = 4$ in our experiments.
With our parameters choice, we obtain the targets in Tab. \ref{tab:targetbudget}.

\begin{table}[htbp]
\centering
\label{tab:targetbudget}
\begin{tabular}{|c|cccc|}
\hline
\backslashbox{  b}{i  } & 1    & 2    & 3    & 4    \\ \hline
4    & 0    & 0    & 0    & 0    \\ \hline
3    & 0.07 & 0.13 & 0.20 & 0.27 \\ \hline
2    & 0.13 & 0.27 & 0.40 & 0.53 \\ \hline
1    & 0.20 & 0.40 & 0.60 & 0.80 \\ \hline
\end{tabular}
\caption{Values of $t^b_i$ for the hyperparameters chosen in our experiments. Every row corresponds to a set of drop predictors associated with the value $b$ of the budget parameter.}
\end{table}

Each row of Tab. \ref{tab:targetbudget} corresponds to a target that will be used to train a set of drop predictor modules. 

\subsection{Point Cloud Classification on ModelNet40}
\label{sec:classif_res}

ModelNet40 consists of 12,311 CAD models distributed across 40 object categories, making it a popular choice for benchmarking point cloud shape classification.
To ensure a fair comparison, we adopted the official split, allocating 9,843 objects for training and 2,468 for evaluation.
We adopted the same sampling strategy described by \citet{qi2017pointnet} to uniformly sample each object to 2048 points.

During training, a random translation in the range $[-0.2, 0.2]$, a scaling in the range $[0.67, 1.5]$, and a random input dropout were applied to augment the input data.
During inference, no augmentation is applied to the data. 

A batch size of 64 was used during training. 
The initial learning rate was set to 0.001, with a cosine annealing scheduler adjusting it at every epoch.

\begin{figure}[!htb]
    \centering
    \includegraphics[width=0.8\textwidth]{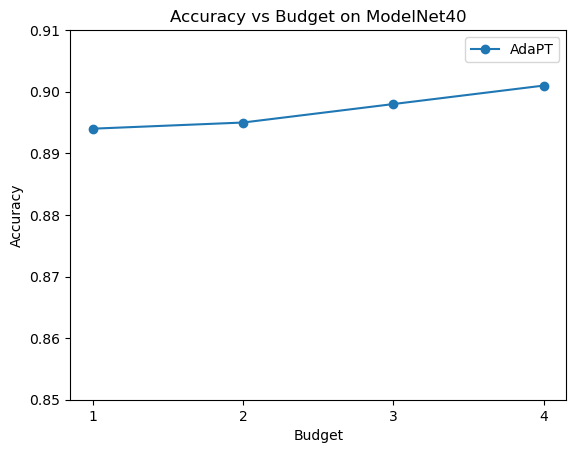}
    \caption{Model performance when selecting different budgets. A stronger budget means more eliminated tokens at each layer. We can see that the model manages to keep a stable performance when the intensity of the budget varies.}
    \label{fig:ablationbudget}
\end{figure}

The experimental results for this classification task are presented in Table \ref{tab:classif}. 
We can see that our model performs similarly to analogous transformer-based models.
More importantly, we can see that our model does not suffer a significant loss of performance when the token drop is introduced. We can also see that the performance is approximatively stable across all the different tested budgets, as shown visually in Fig. \ref{fig:ablationbudget}. We underline again that the results in Fig. \ref{fig:ablationbudget} with the \textit{same} model while varying the budget as an input parameter.
\begin{table}[!htb]
\centering
\label{tab:classif}
\begin{tabular}{c|c}
\hline
Model             & Accuracy \\ \hline
PCT               & 91.0     \\
Point Transformer & 90.8     \\
$\text{AdaPT}_{b = 4}$       & 90.1     \\
$\text{AdaPT}_{b = 3}$       & 89.8     \\ 
$\text{AdaPT}_{b = 2}$       & 89.5      \\
$\text{AdaPT}_{b = 1}$       & 89.4     \\\hline
\end{tabular}
\caption{Shape classification results on the ModelNet40 dataset.}
\end{table}
\newpage
\subsection{Flops count evaluation}

We evaluated the inference FLOPS required by our model while varying the number of initial tokens.\footnote{To evaluate the computational efficiency of our model, we calculate the inference flops using the FlopCountAnalysis method from the fvcore library.}
Even though the regularization term in \eqref{eq:regterm} pushes the model towards the desired number of dropped tokens, this number is not fixed during training.
During inference, the number of tokens to be kept at each layer is deterministic, and the dropped tokens are completely eliminated, making it possible to count the model's flop reliably.

\begin{figure}[htbp]
    \centering
    \includegraphics[width=0.8\textwidth]{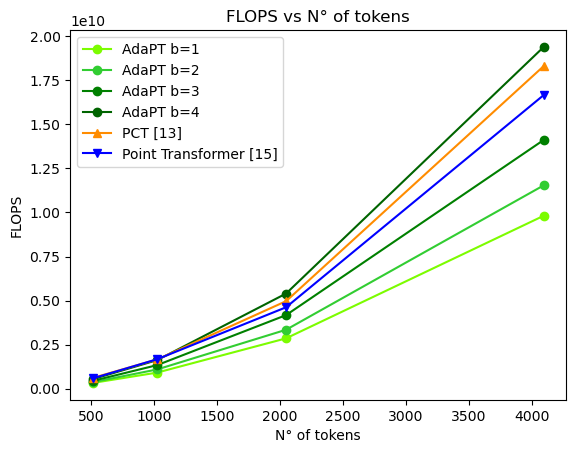}
    \caption{Flops in function of the number of tokens in the transformer part of the model.}
    \label{fig:effplot}
\end{figure}

The results are shown in Figure \ref{fig:effplot}. 
As expected, our model is significantly more efficient when eliminating tokens.

\begin{figure}[!htb]
    \centering
    \includegraphics[width=0.8\textwidth]{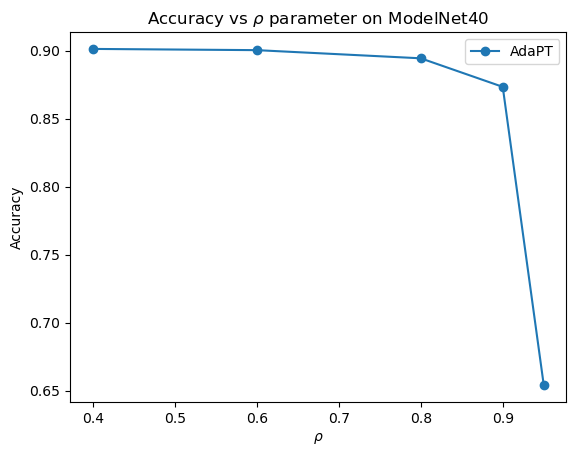}
    \caption{Accuracy of the model in the shape classification task on the ModelNet40 dataset as a function of the $\rho$ hyper-parameter.}
    \label{fig:accvsflops}
\end{figure}

We further investigated the effect of token selection on the model's performance. 
We trained the model with different values of the hyperparameter $\rho$ and observed the accuracy on the ModelNet40 classification task.
We only consider the highest possible value for the budget parameter in these experiments.
The result of these experiments are shown in Figure \ref{fig:accvsflops}. 
As expected, we observe a trade-off between model accuracy and number of dropped tokens. 
We also observe a significant drop in performance when the drop target is set too high, requiring the model to eliminate $95\%$ of the tokens in the last layer.

\newpage
\subsection{Ablation studies}
\label{sec:ablation}

In this section, we present additional experiments aimed at investigating the effectiveness and robustness of our model.
To conduct these experiments, we fix the budget parameter to $b=1$, corresponding to the most severe drop targets.

To test the effectiveness of our sampling strategy, we replace it with alternative, simpler sampling methods.
In particular, we tested two different setups as alternatives to our method of token selection: 
\begin{itemize}
    \item \textbf{Random Selection:} the drop predictor selections are replaced by random samplings of the tokens. This can be seen as a form of token dropout.
    \item \textbf{Farthest Point Sampling:} We perform farthest point sampling on the tokens, in place of the selection made by the drop predictors. We perform this sampling by computing distances with respect to the original points' tridimensional positions.
\end{itemize}
Both performed worse than our sampling strategy. This indicates that our sampling strategy succeeds in selecting more informative tokens than the other two non-learnable sampling methods. 
The results of this test are shown in Table \ref{tab:ablationsampling}.

\begin{table}[htbp]
\begin{tabular}{c|c}
\hline
Sampling strategy & Accuracy \\ \hline
Random            & 79.5     \\
Farthest-Point    & 85.2     \\
Adaptive & 89.4    
\end{tabular}
\centering
\caption{Shape classification results on the ModelNet40 dataset using different sampling methods.}
\label{tab:ablationsampling}
\end{table}

\newpage

\section{Conclusions}
\label{sec:conclusions}

In this paper, we presented Adaptive Point cloud Transformer (AdaPT), a model for efficient point cloud classification.
We employed an end-to-end learnable and task-agnostic sampling mechanism relying on the Gumbel-Softmax distribution to achieve differentiable sampling.
Moreover, we designed a budget mechanism capable of regulating the model's inference FLOPS without retraining. Results on the benchmark dataset ModelNet40 for a point cloud classification task demonstrate the effectivness of the proposed model. 

Future developments of this work include extending the model's task capability to include point cloud segmentation and further investigating the impact of learnable point subsampling on the model's interpretability.

\section*{Declaration of competing interest}

No author associated with this paper has any potential or pertinent conflicts which may be perceived to have impending conflicts with this work.

 \bibliographystyle{elsarticle-num-names} 
 \bibliography{cas-refs}





\end{document}